\documentclass[
]{ceurart}

\usepackage{listings}
\begin{document}

\copyrightyear{2025}
\copyrightclause{Copyright for this paper by its authors.
  Use permitted under Creative Commons License Attribution 4.0
  International (CC BY 4.0).}

\conference{CLEF 2025 Working Notes, 9 -- 12 September 2025, Madrid, Spain}


\title{Overview of PlantCLEF 2025: Multi-Species Plant Identification in Vegetation Quadrat Images}
\title[mode=sub]{Notebook for the LifeCLEF Lab at CLEF 2025}

\tnotemark[1]

\author[1]{Giulio Martellucci}[%
orcid=0009-0008-1804-0920,
email=giulio.martellucci@inrae.fr
]
\author[2,3]{Herv\'e Go\"eau}[%
orcid=0000-0003-3296-3795,
email=herve.goeau@cirad.fr
]
\author[2]{Pierre Bonnet}[%
orcid=0000-0002-2828-4389,
email=pierre.bonnet@cirad.fr
]
\author[1]{Fabrice Vinatier}[%
orcid=0000-0003-3693-4422,
email=fabrice.vinatier@inrae.fr
]
\author[3]{Alexis Joly}[%
orcid=0000-0002-2161-9940,
email=alexis.joly@inria.fr
]
\address[1]{LISAH, Univ Montpellier, INRAE, IRD, Montpellier, France}
\address[2]{CIRAD, UMR AMAP, Montpellier, Occitanie, France}
\address[3]{Inria, LIRMM, Univ Montpellier, CNRS, Montpellier, France}


\begin{abstract}
Quadrat images are essential for ecological studies, as they enable standardized sampling, the assessment of plant biodiversity, long-term monitoring, and large-scale field campaigns. These images typically cover an area of fifty centimetres or one square meter, and botanists carefully identify all the species present. Integrating AI could help specialists accelerate their inventories and expand the spatial coverage of ecological studies. To assess progress in this area, the PlantCLEF 2025 challenge relies on a new test set of 2,105 high-resolution multi-label images annotated by experts and covering around 400 species. It also provides a large training set of 1.4 million individual plant images, along with vision transformer models pre-trained on this data. The task is formulated as a (weakly labelled) multi-label classification problem, where the goal is to predict all species present in a quadrat image using single-label training data. This paper provides a detailed description of the data, the evaluation methodology, the methods and models used by participants, and the results achieved.
\end{abstract}


\maketitle

\section{Introduction}

Vegetation plot inventories are essential for ecological studies, as they enable standardized sampling, biodiversity assessment, long-term monitoring, and large-scale remote surveys. They provide valuable data for understanding ecosystem structure and dynamics, guiding biodiversity conservation strategies, and supporting evidence-based environmental decision-making. 

Vegetation plots - commonly referred to as "quadrats" - are typically 0.5 x 0.5 meters in size. Within each quadrat, botanists meticulously identify all present plant species and quantify their abundance using indicators such as biomass, cover percentage, and presence-based qualification factors derived from photographic analysis.

\begin{figure}
    \centering
    \begin{tabular}{p{0.54\linewidth}|p{0.42\linewidth}}
         \includegraphics[width=\linewidth, trim={14.8cm 0 0cm 0},clip]{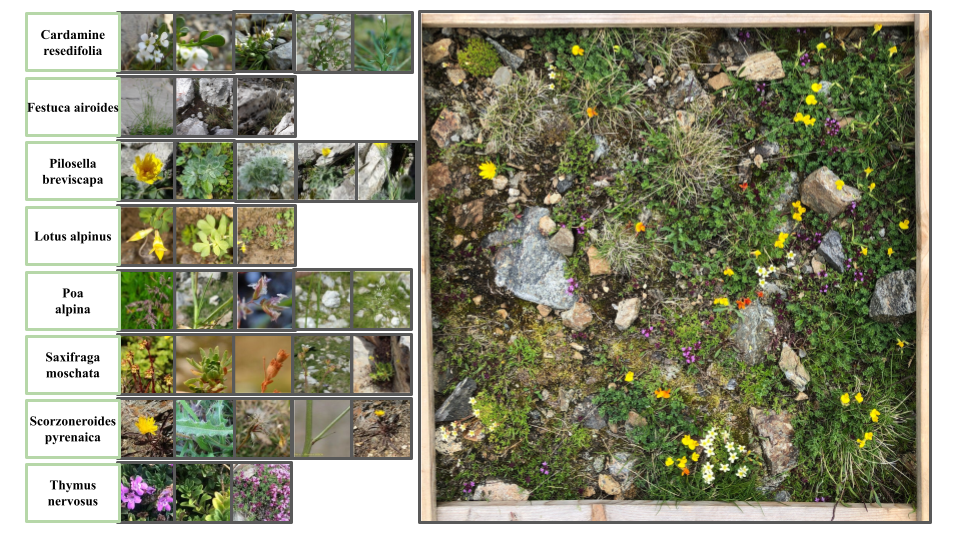} 
         & \includegraphics[width=\linewidth, trim={0 0 19.2cm 0},clip]{Figures/PlantCLEF_2024_one_slide.png} \\ 
         (a) Vegetative quadrat (test)
         &
         (b) Individual plant images (training)\\
    \end{tabular}    
    \caption{Illustration of the visual discrepancy between (a) the test set composed exclusively of vertical top views of quadrats with potentially many plant species and (b) the training set based on images of individual plants focusing mostly on organs (flowers, fruits, leaves, stems) with various angles and an wide range of shooting conditions due to the collaborative nature of the original dataset.}
    \label{fig:PlantCLEF2024discrepancy}
\end{figure}

AI could significantly improve the efficiency of this process, increasing the frequency and spatial resolution of ecological surveys. It could also empower non-expert citizen scientists to contribute to monitoring programs. 
Existing plant identification applications, such as Pl@ntNet and iNaturalist, can already automatically identify specimens in a quadrat by photographing them one by one. However, a much more efficient approach would involve identifying all species within a quadrat  from a single high-resolution image (see Figure \ref{fig:PlantCLEF2024discrepancy}(a)). Developing AI models that can solve this multi-label classification task remains a major challenge. 

Ideally, training models for this task would require a large annotated dataset of quadrat images, each exhaustively labelled with all the visible species. Yet, producing such a dataset is prohibitively expensive and time-consuming as it requires considerable expert effort to cover the thousands of species of an entire flora. In contrast, millions of labelled images of individual plants are available through platforms like Pl@ntNet, which allow for training highly effective species-level classifiers.\cite{plantclef2022,plantclef2023}. 

To bridge this gap, the previous edition of the PlantCLEF challenge \cite{plantclef2024overview} explores the task of predicting species in quadrat images using only individual plant images for training.  Given the difficulty of the task and the limited floristic diversity in the previous test set, the 2025 challenge addresses the same problem by introducing a more diverse and representative selection of quadrat images for evaluation. 

As shown in Figure \ref{fig:PlantCLEF2024discrepancy}, the main difficulty lies in the domain shift between training and test data. While the test data consist of high-resolution images of vegetation plots containing multiple species, the training set mainly consists of single-label images of individual plants or plants organs, sourced from the Pl@ntNet collaborative platform \cite{affouard2017pl}. In addition to viewpoint differences and shooting conditions, variations in phenological stage (i.e., the plant’s developmental phase, such as flowering, fruiting, seedling, or senescence) further increase data disparity. Unlike individual plant images, which often show close-ups of flowers, quadrat images are collected across seasons and years, with no control over the plants' phenological stage: some may be flowering, some fruiting, some in early growth or decline, and some may even be affected by disease. 

\section{Dataset}

\subsection{Training dataset}

The training dataset includes observations of individual plants, similar to those used in previous editions of PlantCLEF prior to 2024. More precisely, it is a subset of the Pl@ntNet training data that focuses on South-Western Europe and covers 7,806 plant species. It contains approximately 1.4 million images, some of which have been aggregated from the GBIF platform to provide labels for the less-illustrated species. Links to the original images are shared with the participants. The images have a relatively high resolution, with a minimum of 800 pixels on the shortest side or a maximum of 800 pixels on the longest side, allowing the use of classification models capable of handling large resolution inputs, which may reduce the difficulty of predicting small plants in images of  vegetation plots. The images are first organized into species-specific subfolders and split into predefined train, validation, and test sets to simplify data handling, save time, and ease model training setup.

\begin{table}
    \caption{Training dataset statistics. It was provided to participants in the form of three sub-directories, subdividing the data into “Train,” “Val,” and “Test” sets to facilitate the training of individual plant identification models. It is important not to confuse this “Test” set, dedicated to evaluating the performance of potential individual plant identification models, with the challenge test set, which contains large multi-species images.}
    \centering
        \begin{tabular}{cccccc}
            \toprule
            Dataset (with predefined splits) & Images & Observations & Species & Genera & Families \\
            \hline
            All & 1,408,033 & 1,151,904 & 7,806 & 1,446 & 181 \\
            \midrule
            Train & 1,308,899 & 1,052,927 & 7,806 & 1,446 & 181 \\
            Val & 51,194 & 51,045 & 6,670 & 1,415 & 181 \\
            Test & 47,940 & 47,932 & 5,912 & 1,375 & 181 \\            
            \bottomrule
            \end{tabular}
    \label{tab:stats}
\end{table}

\subsection{Complementary training dataset: unlabelled high-resolution pseudo-quadrat images}

To facilitate model adaptation to the multi-species quadrat images used for evaluation, a supplementary training dataset has been provided. Although annotated multi-species quadrat images would be ideal for supervised learning, such data remains scarce. As an alternative, this dataset consists of high-resolution vegetation images framed similarly to quadrats but not strictly limited to a 50 x 50cm area, and without species-level annotations.

Although these images are unlabelled, they can be used in self-supervised or weakly supervised learning strategies to pre-train models within a domain closer to the target data. This could reduce the domain shift between the single-species training data and the complex multi-species test images.

The dataset is derived from the LUCAS Cover Photos 2006–2018 archive \cite{lucas_cover}. A dedicated model was trained to automatically select relevant pseudo-quadrat views, resulting in a curated subset of 212,782 images.

\subsection{Test dataset}
 
The test set is a compilation of several quadrat image datasets collected across various floristic contexts (including Pyrenean, Mediterranean, and South-Western European temperate floras) and covering a wide range of environments, from remote mountain areas and protected natural reserves to highly heavily human-influenced sites such as popular beaches and field margins. 

\begin{figure}
    \centering
    \begin{tabular}{p{0.31\linewidth}p{0.31\linewidth}p{0.31\linewidth}}
        \includegraphics[width=\linewidth,height=\linewidth]{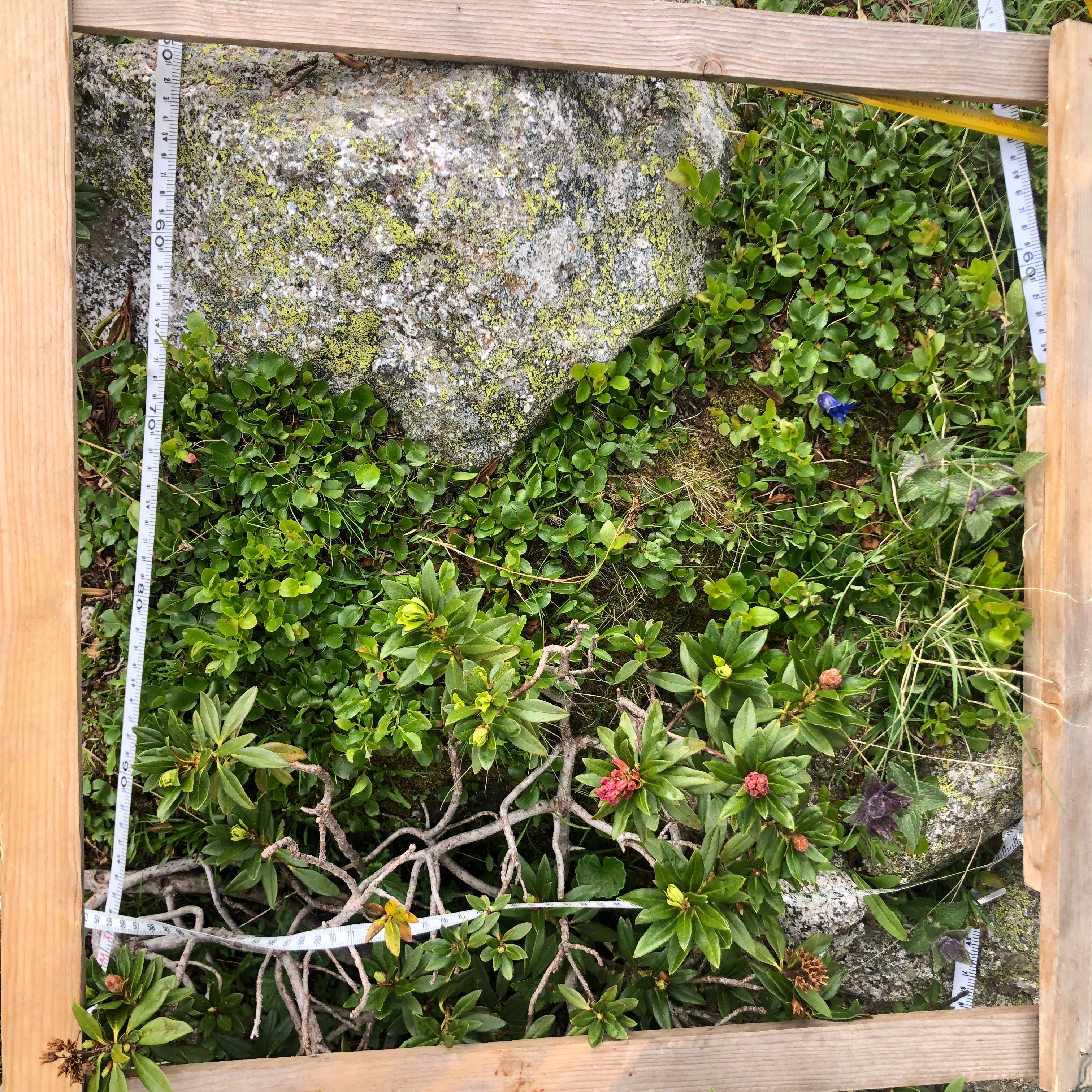} &
        \includegraphics[width=\linewidth,height=\linewidth]{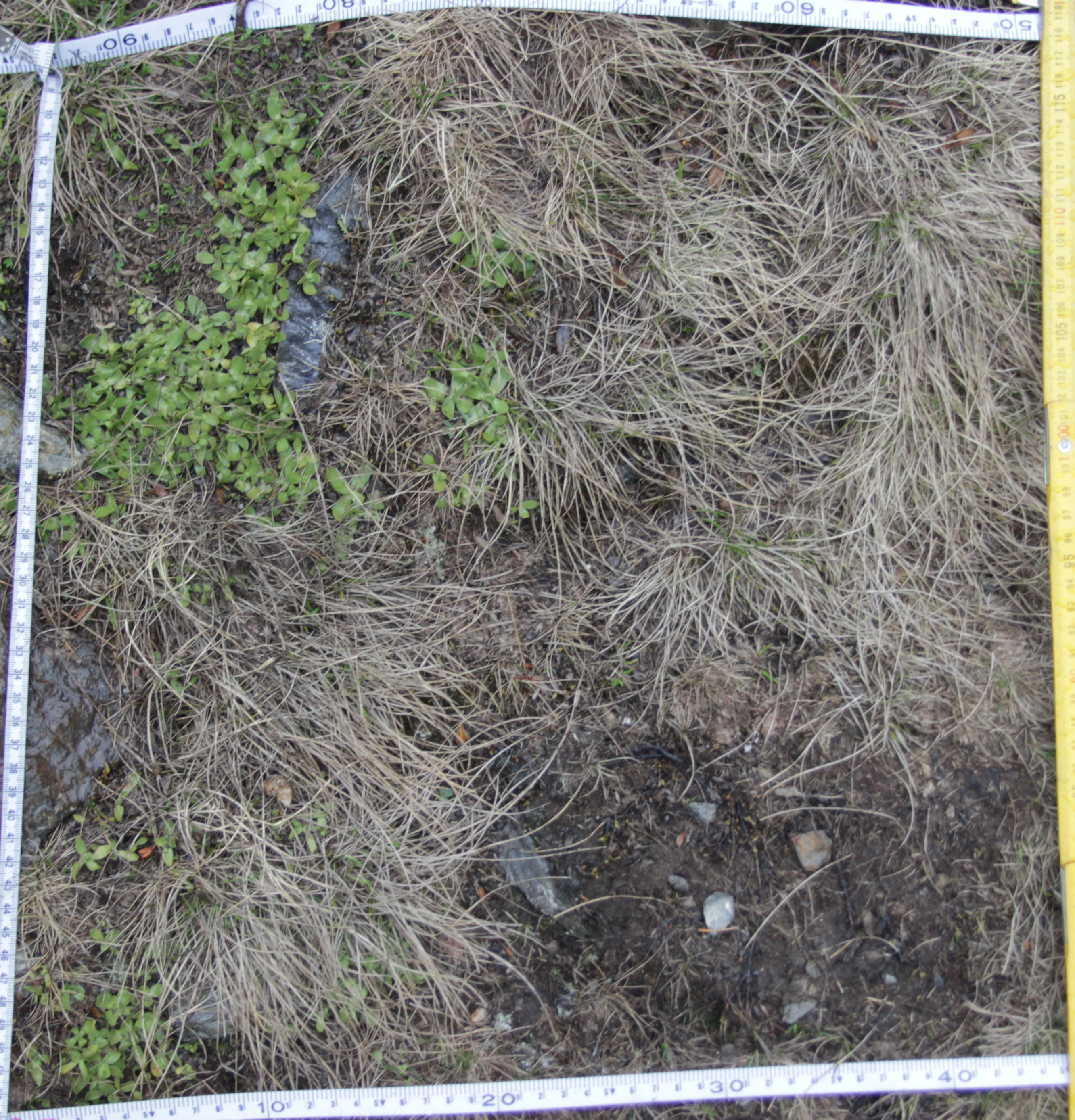} &
        \includegraphics[width=\linewidth,height=\linewidth]{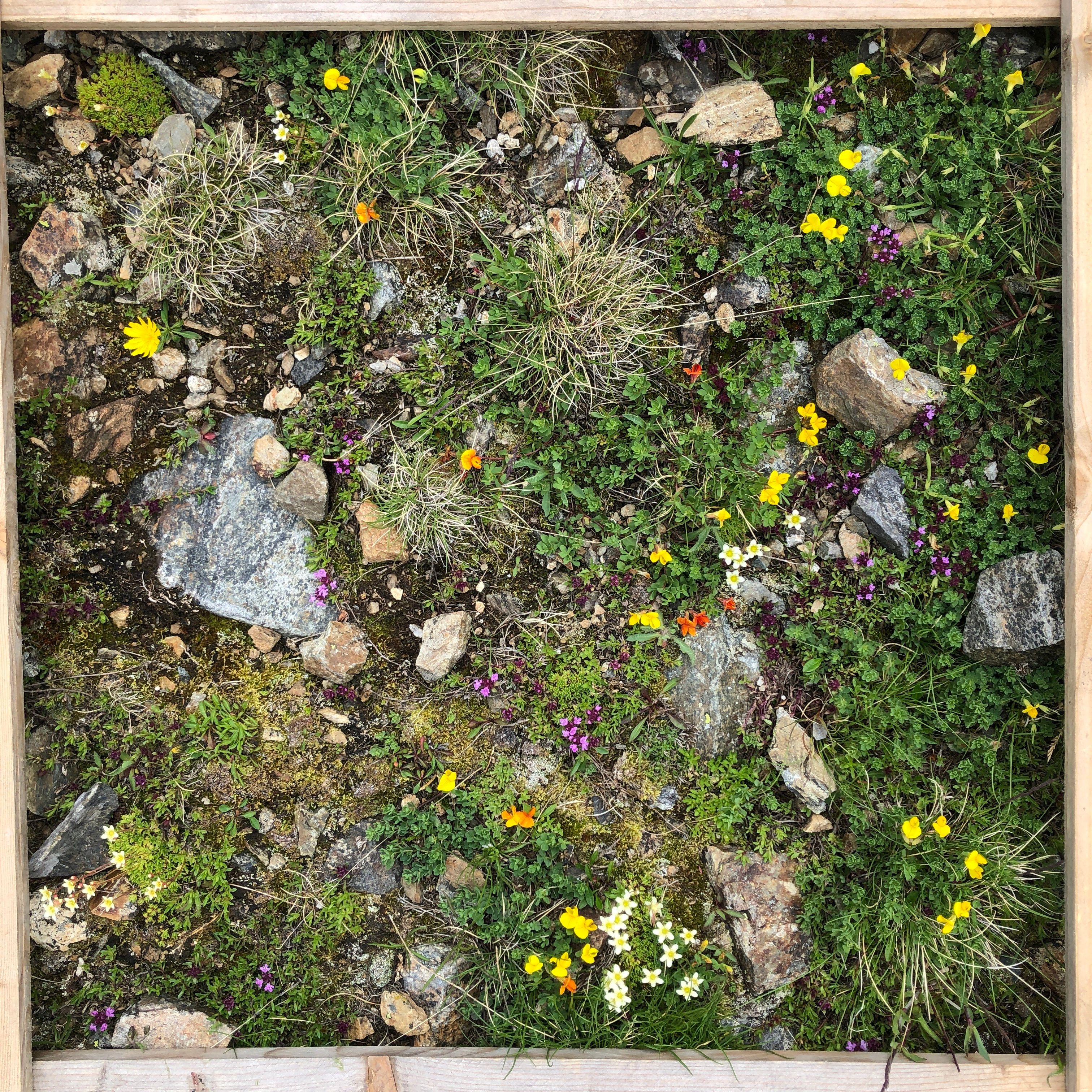} \\
        \includegraphics[width=\linewidth,height=\linewidth]{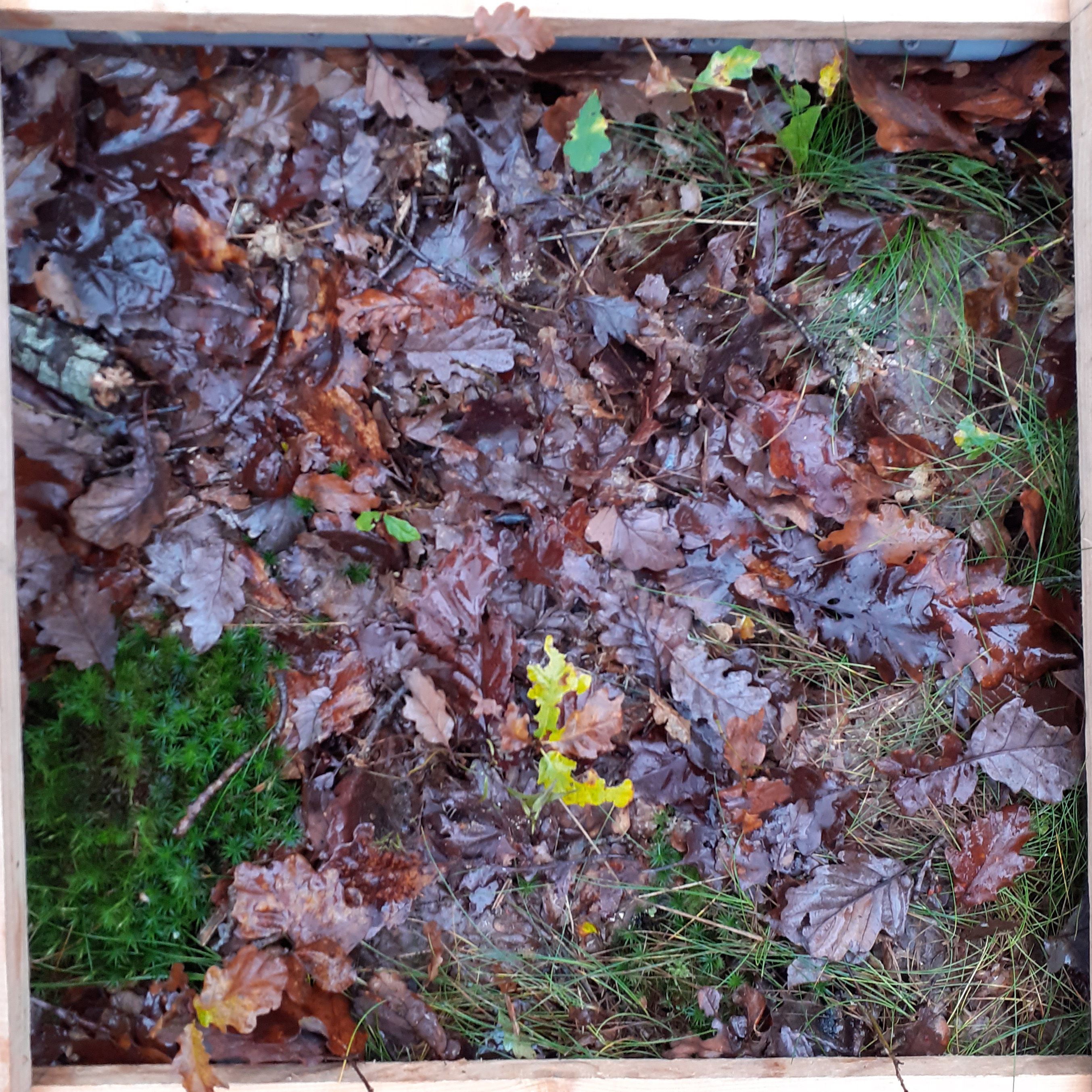} &  
        \includegraphics[width=\linewidth,height=\linewidth]{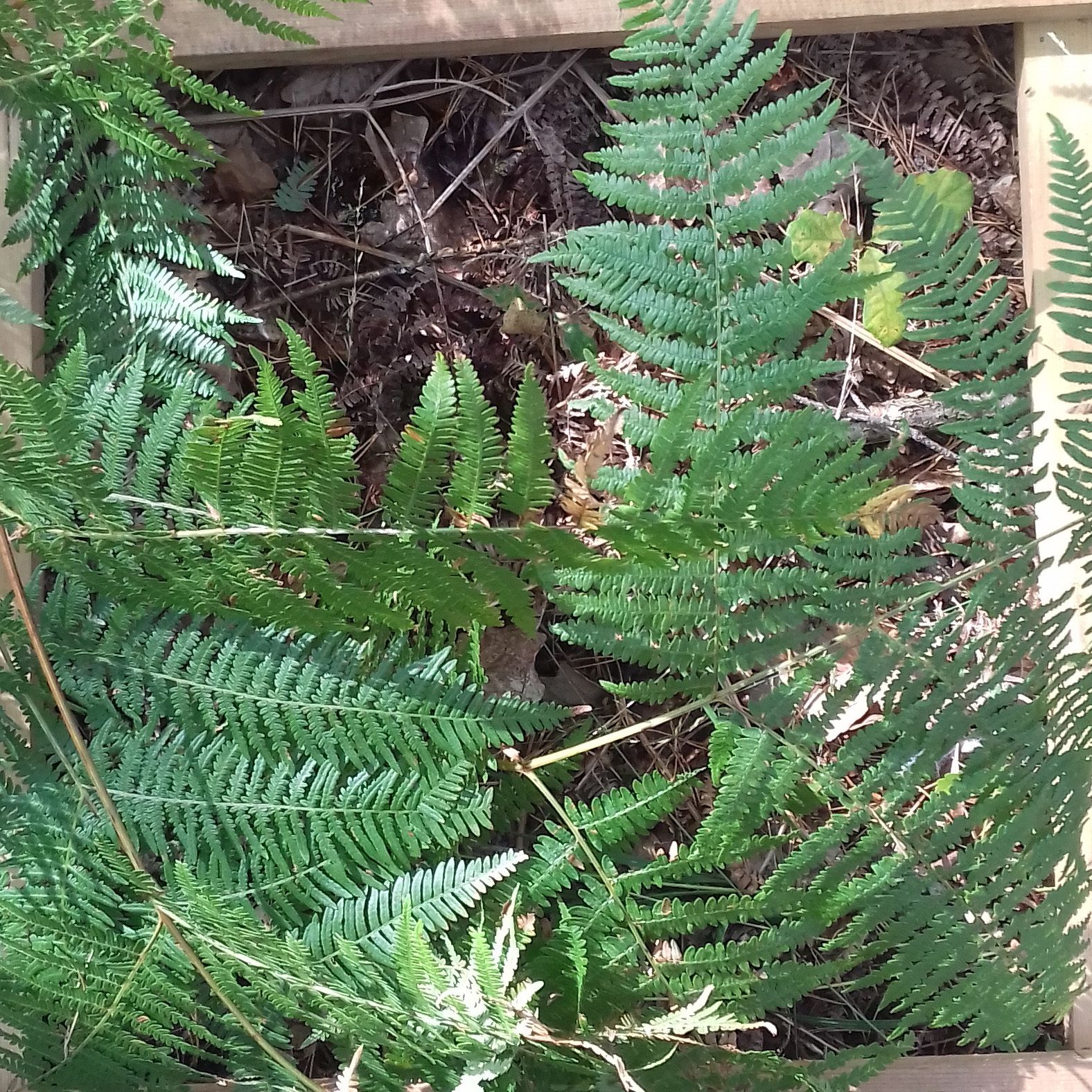} &
        \includegraphics[width=\linewidth,height=\linewidth]{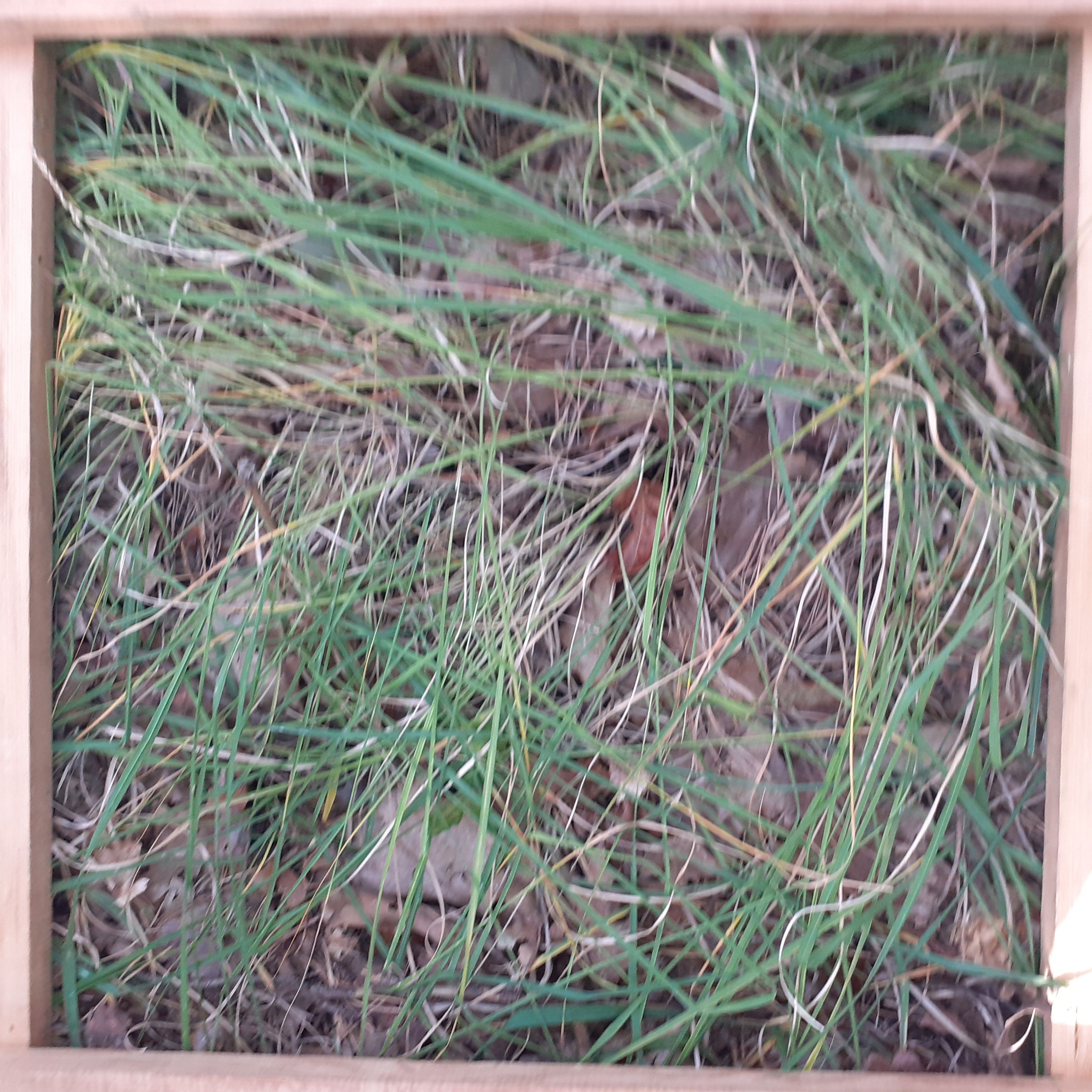} \\
        \includegraphics[width=\linewidth,height=\linewidth]{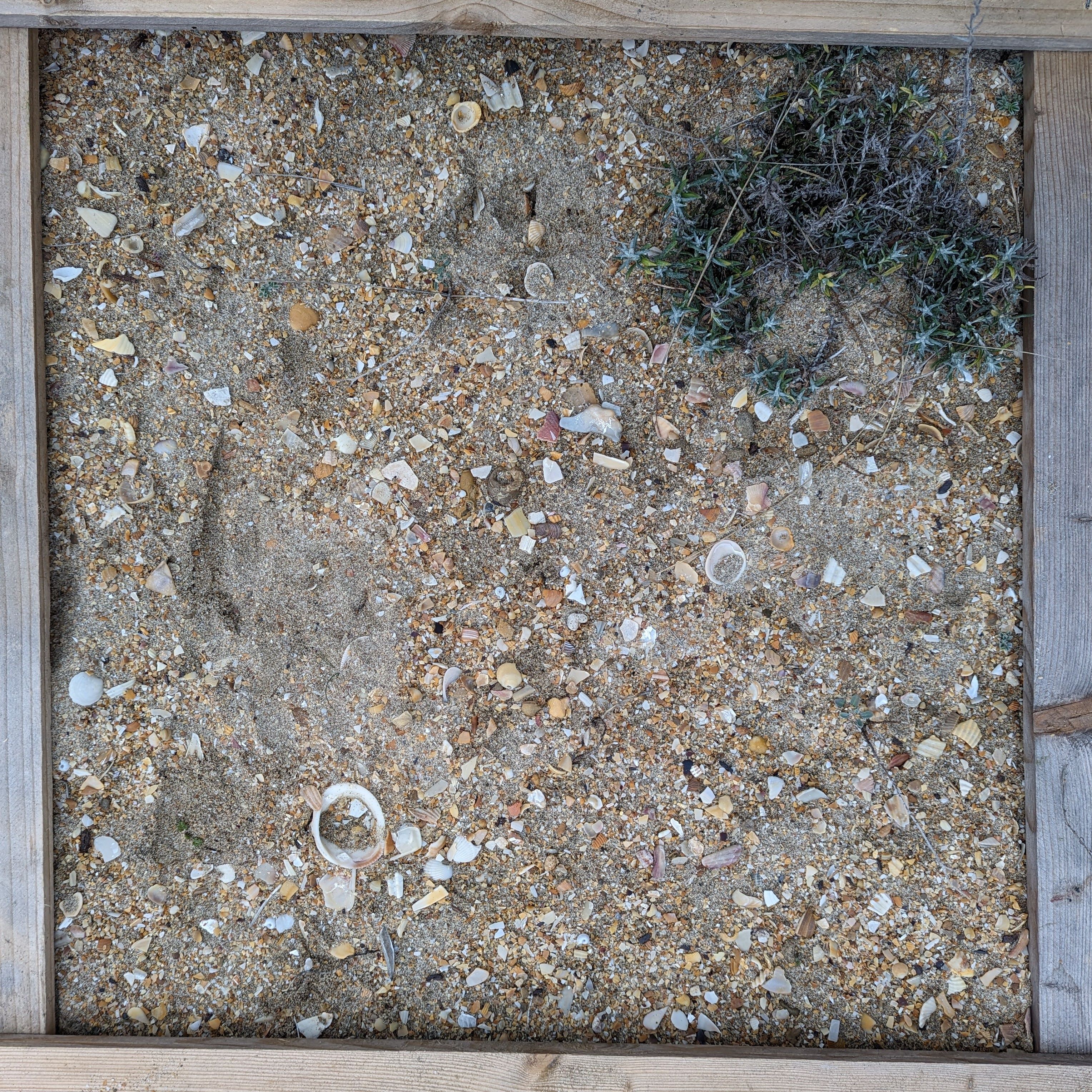} &
        \includegraphics[width=\linewidth,height=\linewidth]{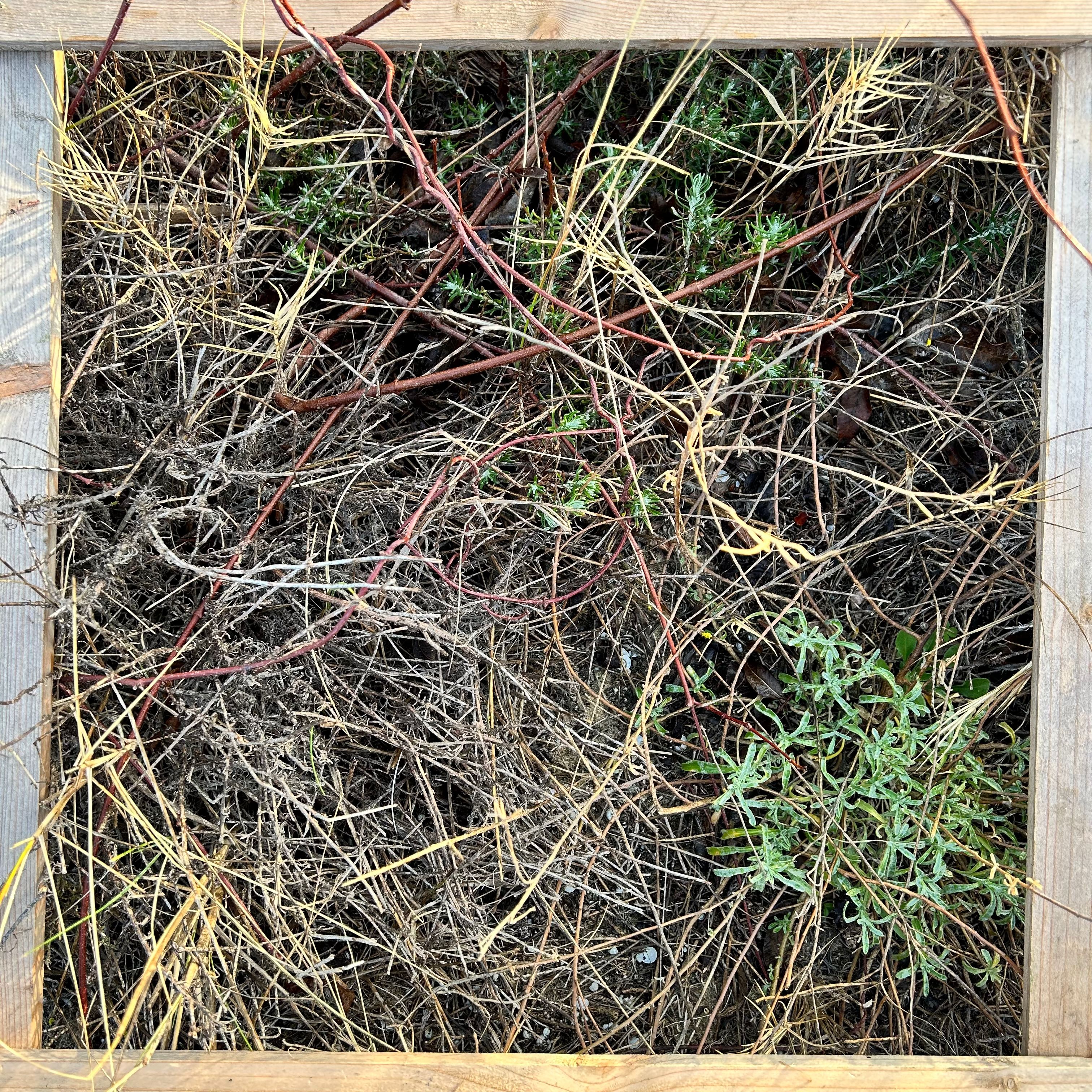} &
        \includegraphics[width=\linewidth,height=\linewidth]{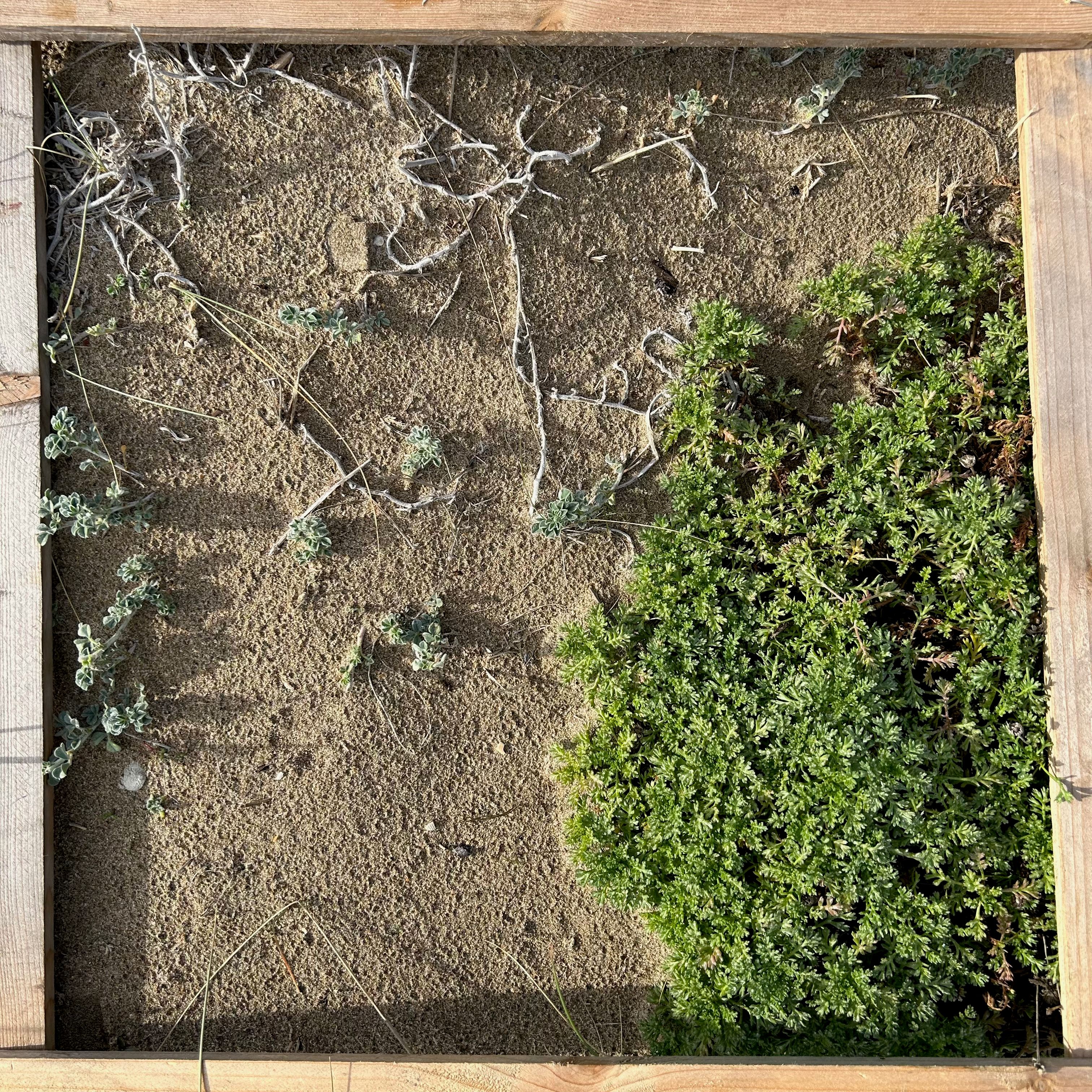} \\
    \end{tabular}    
    \caption{Illustration of the diversity of quadrat images in the test set.}
    \label{fig:morequadrats}
\end{figure}

These datasets are all produced by experts and consist of a total of 2,105 high-resolution images. The shooting protocol can vary significantly from one context to another. For example, the use of wooden frames or measuring tape to delimit the plot or not, the angle of view being more or less perpendicular to the ground due to the slope of the photographed site (see Figure \ref{fig:morequadrats}). Additionally, image quality may vary depending on the weather, resulting in more or less pronounced shadows, blurry areas, and so on.

\subsection{Pre-trained models}
For participants who may have difficulty finding the computational power necessary to train a plant image identification model on such a large volume of data, or to enable direct work on top of a backbone model for image embedding extraction for instance, two pre-trained models are shared through Zenodo \cite{plantclef2024pretrainedmodel}. Both are based on a state-of-the-art Vision Transformer (ViT) architecture initially pre-trained with the DinoV2 Self-Supervised Learning (SSL) approach \cite{oquab2023dinov2, darcet2024vision} and then fine-tuned on the challenge training data.

The vit\_base\_patch14\_reg4\_dinov2\_lvd142m\_onlyclassifier model relies on the original SSL pre-trained weights from the LVD-142M dataset using the DinoV2 SSL method, where only the classification head was trained in a conventional supervised manner, with the backbone being frozen. The second model named vit\_base\_patch14\_reg4\_dinov2\_lvd142m\_onlyclassifier\_then\_all fine-tuned the backbone weights in a classical supervised manner with the classifier head, resulting in a backbone that no longer aligns with an SSL pre-trained model. 

Both models were pre-trained via the timm library \cite{rw2019timm}. Following the naming convention adopted in this library, the second model vit\_base\_patch14\_reg4\_dinov2\_lvd142m\_onlyclassifier\_then\_all could be named vit\_base\_patch14\_reg4\_dinov2\_lvd142m\_ft\_pc24, expressing the fact that the initial DinoV2 ViT pre-trained model is subsequently fine-tuned (ft) on the PlantCLEF2024 dataset in a usual supervised way (all layers are fine-tuned). The vit\_base\_patch14\_reg4\_dinov2\_lvd142m corresponds to the backbone only, without the head and represents the original pre-trained model with DinoV2 on the huge generalist LVD142M dataset of 142 million images. For ease of reading, we will subsequently refer to these two models as ViTD2PC24OC and ViTD2PC24All for vit\_base\_patch14\_reg4\_dinov2\_lvd142m\_onlyclassifier and vit\_base\_patch14\_reg4\_dinov2\_lvd142m\_onlyclassifier\_then\_all, respectively. ViTD2 will stand for the original pre-trained backbone vit\_base\_patch14\_dinov2\_lvd142m but without the four registers features.

The training set was split into three sub-directories, (sub) training, validation, and test sets at the observation level (i.e., all images from a given observation are assigned to the same split), in order to facilitate the training of individual plant identification models by both the organizers and the participants (see Table \ref{tab:stats} for the statistics). It is important not to confuse this test set dedicated to individual plant identification with the challenge test set, which contains high-resolution multi-species images.

The two pre-trained models were both trained on an A100 octo-GPU node server using the Timm library (version 0.9.16) under torch (version 2.2.1+cu121). The ViTD2PC24OC model, with only the classifier head being fine-tuned, was trained for approximately 17 hours over 92 epochs with a batch size of 1280 per GPU and a high learning rate of 0.01 (hyperparameters for data augmentation techniques and other features are detailed in the Zenodo package \cite{plantclef2024pretrainedmodel}). The second model, ViTD2PC24All, was initialized with the weights of the previously trained ViTD2PC24OC model and was then fully fine-tuned (all layers) for approximately 36 hours over 92 epochs with a significantly lower batch size per GPU (144) and learning rate (0.00008). Both models were trained with a cross-entropy loss since the objective of the training is to predict a single species per image.

Table \ref{tab:pretrained_perf} shows the performances of the two pre-trained models on the subset excluded from the training and used as a "Test" set for evaluating performances and model's generalization capacity to identify on individual plants (note again that this "Test" set of individual plants should not be confused with the challenge test set of vegetation quadrat images).
\begin{table}
    \caption{Top-1 and top-5 accuracies of the pre-trained model on single-plant images (not quadrat images)}
    \centering
        \begin{tabular}{llcc}
            \toprule
            Pre-trained model & Short name & Top1 & Top5 \\
            \midrule
            vit\_base\_patch14\_reg4\_dinov2\_lvd142m\_onlyclassifier               & ViTD2PC24OC & 63.69 & 83.88 \\
            vit\_base\_patch14\_reg4\_dinov2\_lvd142m\_onlyclassifier\_then\_all    & ViTD2PC24All& 75.91 & 92.82 \\
            \bottomrule
        \end{tabular}
    \label{tab:pretrained_perf}
\end{table}

 
\section{Task description}

The aim of the challenge is to detect the presence of every plant species in each high-resolution vegetation plot image, among more than 7,800 species Plots are generally 50 x 50cm in size and that it is uncommon to observe dozens of species simultaneously. Formally, this is a weakly supervised multi-label classification task, for which only single-label training data is available, and that involves a significant distribution shift.

\subsection{Metric}

The metric chosen to evaluate the methods of the participants is the F1 score, designed to balance recall and precision, avoiding excessive predictions that would reduce precision, while also penalizing species omissions that would lower recall. Among the several F1 score variations such as the macro-averaged per class or the micro-averaged, the macro-averaged per sample was selected. This metric computes the F1 score for each test image (i.e. vegetation quadrat) individually, and then averages the scores across all test images.

\begin{figure}
    \centering
    \includegraphics[width=1.0\linewidth]{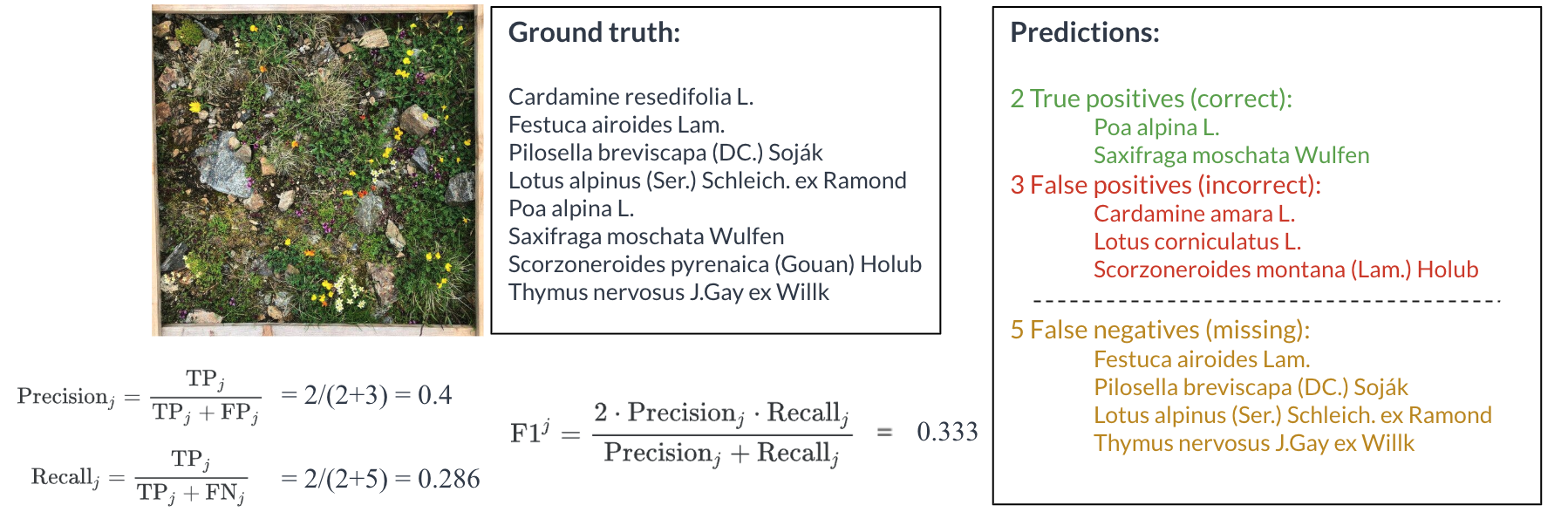}
    \caption{Illustration of per-sample F1 score calculation for a quadrat image.}
    \label{fig:PlantCLEF2025F1PerSampleExample}
\end{figure}

More formally, the macro-averaged per sample F1 score is: 
\begin{equation}
\text{F1}_{\text{macro-averaged per sample}} = \frac{1}{N} \sum_{i=1}^{N} F1_i
\end{equation}
where:
\begin{itemize}
    \item $N$ is the number of test images.
    \item $F1_i$ is the F1 score for image $i$, calculated as:
    \begin{equation}
    F1_i = \frac{2 \cdot \text{Precision}_i \cdot \text{Recall}_i}{\text{Precision}_i + \text{Recall}_i}
    \end{equation}
    \item $\text{Precision}_i$ and $\text{Recall}_i$ are the precision and recall for test image $i$, defined as:
    \begin{align*}
    \text{Precision}_i &= \frac{\text{TP}_i}{\text{TP}_i + \text{FP}_i} \\
    \text{Recall}_i &= \frac{\text{TP}_i}{\text{TP}_i + \text{FN}_i}
    \end{align*}
    \item where for test image $i$:
    \begin{itemize}
        \item $\text{TP}_i$: true positives (correctly predicted species).
        \item $\text{FP}_i$: false positives (species predicted but not present).
        \item $\text{FN}_i$: false negatives (species present but not predicted).
    \end{itemize}
\end{itemize}
Figure~\ref{fig:PlantCLEF2025F1PerSampleExample} provides a detailed example of an F1 per sample score computed on a quadrat test image.

However, in ecological studies, quadrats are usually sampled along "transects", which are spatially and temporally structured to capture biodiversity patterns over time and space. To avoid bias due to irregular sampling density or oversampling of certain areas, the final evaluation metric incorporates a second level of aggregation: F1 scores are first averaged across the quadrats within the same transect sampled at different time, and then across all transects in the test set. The final macro-averaged per sample F1 score is thus computed as:
\begin{equation}
F1_{\text{avg-transect}} = \frac{1}{T} \sum_{k=1}^{T} \left( \frac{1}{Q_k} \sum_{j=1}^{Q_k} F1^{(k)}_j \right)
\end{equation}

where:
\begin{itemize}
    \item $T$ is the total number of transects in the test set. Note that a transect surveyed several times counts as only one.
    \item $Q_k$ is the number of quadrat images in transect $k$.
    \item $F1^{(k)}_j$ is the macro-averaged per-sample F1 score for the $j^{\text{th}}$ quadrat image in transect $k$.
\end{itemize}

This hierarchical averaging ensures that the final score is not dominated by a single transect due to a larger number of quadrats nor by a transect that has been surveyed several times. It also reflects the ecological structure of the sampling design, making it more aligned with real-world biodiversity monitoring practices.

\subsection{Rules about metadata and additional data}

The use of the metadata (licenses, exif) was authorised provided that, for each run using metadata, an equivalent run using only the visual information is submitted in order to assess the raw contribution of a purely visual analysis. 

The use of additional data is permitted, on the condition that an equivalent run with only the data provided is submitted to enable more accurate and fair comparisons.

\subsection{Challenge platform}

The PlantCLEF2025 challenge was hosted on the \href{https://www.kaggle.com/competitions/plantclef-2025/}{Kaggle platform}, providing an opportunity for researchers and enthusiasts to contribute to the development of plant recognition in such an original context of plot analysis.

\section{Participants and methods}

Out of the 540 initial entrants who expressed interest in participating, 55 participants grouped into 38 teams ultimately submitted 659 runs. Ten teams produced and shared working notes.

Before comparing the ten teams’ approaches, it is worth noting that all but two of them adopted a tiling inference pipeline, generating predictions on sub-images (“tiles”) extracted from high-resolution test images. Since the training set focuses on individual plants, sub-sampling test images to isolate single-species regions is a logical strategy. Most teams used the pre-trained ViTD2PC24All model, while some trained additional models. The tiling parameters varied: initial image resolution, tile size, step size, and sometimes a border offset to exclude elements such as wooden frames or measuring tapes.

Two teams diverged from the common tiling approach by explicitly detecting regions of interest to be classified. One team relied on a conventional object detector (YOLO), while the other team used attention maps from a dedicated trained model to identify informative areas, bypassing any systematic tiling strategy.

Finally, as each tile or region has its own set of predictions, various strategies were implemented to limit (e.g., top-k), threshold (e.g., based on probability), and aggregate the predictions at the quadrat image level.

The following subsections detail the distinguishing features of these approaches.

\subsection{TheHeartOfNoise \#Rust \#Candle, Vincent Espitalier, Independant, France, 43 runs, \textbf{Rank: 1} \cite{espitalier2025preprocessing}}

This participant, who achieved the highest score in the competition, focused on the impact of image pre-processing. A key contribution is the use of careful configuration of JPEG recompression, tuning both quality and YCbCr subsampling. This significantly improves F1 scores by acting as a form of regularization and reducing the distribution gap between training and test data. The study compares various tiling strategies with high-resolution single-shot inference methods, including variants with modified attention mechanisms. The entire pipeline is implemented in Rust, a language rarely used in this context. The results demonstrate that seemingly minor preprocessing choices can lead to substantial performance gains.

\subsection{DS@GT-LifeCLEF, Georgia Institute of Technology, USA, 29 runs, \textbf{Rank: 2} \cite{gustineli2025transformers}}

This team used the Vision Transformer model provided by the organisers, applying a 4×4 tiling strategy and aggregating predictions by species frequency across tiles. To mitigate domain shift and class imbalance, two alternative strategies were explored: (1) geolocation-based, which filters predictions to species observed within the test region, and (2) region-specific Bayesian reweighting, which used priors derived from visual clustering. The latter involves dimensionality reduction with PaCMAP \cite{wang2021understanding} followed by K-Means clustering of test image embeddings to define ecologically coherent groups, each with adjusted species probabilities. Although segmentation-based approaches using Grounded SAM were explored to isolate leaves and flowers, these were not included in the final submission. The best result was obtained by combining tiling with cluster-based prior reweighting.

\subsection{Chlorophyll Crew, University of Hamburg, Germany, 77 runs, \textbf{Rank: 3} \cite{herasimchyk2025metadata}}

Ranked 3rd, this team based their approach on multiple classification heads (species, genus, and family) built upon the provided pre-trained model. They assumed that leveraging the taxonomic hierarchy would improve prediction quality. Several additional techniques were also employed, such as multi-scale tiling to capture plants at different sizes, dynamic logit thresholding based on the average prediction length, ensemble methods, top-N filtering, image cropping to remove non-plant artifacts, and metadata-based inference adjustments.

\subsection{webmaking, Moscow Pedagogical State University, Russia, 202 runs, \textbf{Rank: 4}\cite{webmaking2025plant}}

The webmaking team developed and combined various complementary techniques through a post-processing pipeline. Using the Vision Transformer model provided by the organizers, their approach integrates multi-scale tiling with test-time augmentation, zero-shot segmentation via GroundingDINO and SAM to down-weight non-vegetal regions, seasonal filtering using GBIF occurrence data, ecological score adjustments based on niche similarity computed from Ellenberg indicator values (EIVE) \cite{dengler2023eive}, and cross-year aggregation.

\subsection{ADAM, University of Feira de Santana, Brazil, 16 runs, \textbf{Rank: 5}\cite{dourado2025zeroshot}}

The ADAM team introduced an original segmentation-assisted framework for multi-label species classification, that avoids standard tiling strategies. Using the pre-trained Vision Transformer model provided by the organisers, their method leverages class prototypes derived from K-means clustering of DINOv2 embeddings computed on single-species training images. A lightweight Vision Transformer is trained to reconstruct these prototypes from DINOv2 embeddings of test images. This proxy objective guides the model's attention toward plant-relevant regions, which are then used for classification. Contextual aggregation around high-attention zones was found to outperform direct patch-wise predictions.

\subsection{NEUON\_AI, NEUON AI \& Swinburne University of Technology Sarawak, Malaysia, 9 runs, \textbf{Rank: 7}\cite{ishrat2025posthoc}}

The NEUON\_AI team focused on post-hoc aggregation of patch-wise predictions rather than modifying the underlying model. Using the pre-trained DINOv2 Vision Transformer provided by the organisers, they explored several self-supervised learning (SSL) pretext tasks — such as color-based adaptations — to address domain shifts caused by blur, occlusion, and seasonality. These approaches underperformed, likely due to the limited training data available. The team then refined a Bayesian Model Averaging (BMA) strategy previously used in 2024 \cite{chulif2024patch}, incorporating patch-level confidence metrics such as variance, entropy, and a custom vegetation percentage estimator.

\subsection{VerbaNexAI, Universidad Tecnológica de Bolívar, Colombia, 7 runs, \textbf{Rank: 13}\cite{menco2025patch}}

The VerbaNexAI team explored a patch-based classification strategy using the pre-trained Vision Transformer model provided by the organisers, applied to high-resolution vegetation plot images. Test images were systematically divided into various patch configurations (1×1, 4×2, 3×3, 14×7, 16×16), and predictions were performed independently on each patch and then aggregated using simple thresholding.

\subsection{I2C-UHU-PERSEUS,  University of Huelva, Spain, 6 runs, \textbf{Rank: 28}\cite{tejion2025i2c}}

The I2C-UHU-PERSEUS team proposed a hybrid pipeline combining object detection with multi-label classification for plant species identification in high-resolution quadrat images. Using YOLOv11 to localize relevant regions and a custom InceptionV3 model for classification, their approach aimed to leverage region-level focus prior to label prediction. To mitigate extreme class imbalance, they applied random undersampling by discarding under-represented species and employed data augmentation.

\subsection{Shadow Park, Sri Sivasubramaniya Nadar College of Engineering, India, 10 runs, \textbf{Rank: 32} \cite{jayasree2025tiling}}

The Shadow Spark team addressed the challenge with a tiling-based pipeline using the DINOv2 Vision Transformer. They extracted tile embeddings from both the base and organizer-provided fine-tuned models. They applied a 3×3 tiling strategy, used kNN classifiers on each tile, and aggregated the top-5 species per tile into final predictions.

\section{Results}
\begin{figure}
    \centering
    \includegraphics[width=1.0\linewidth]{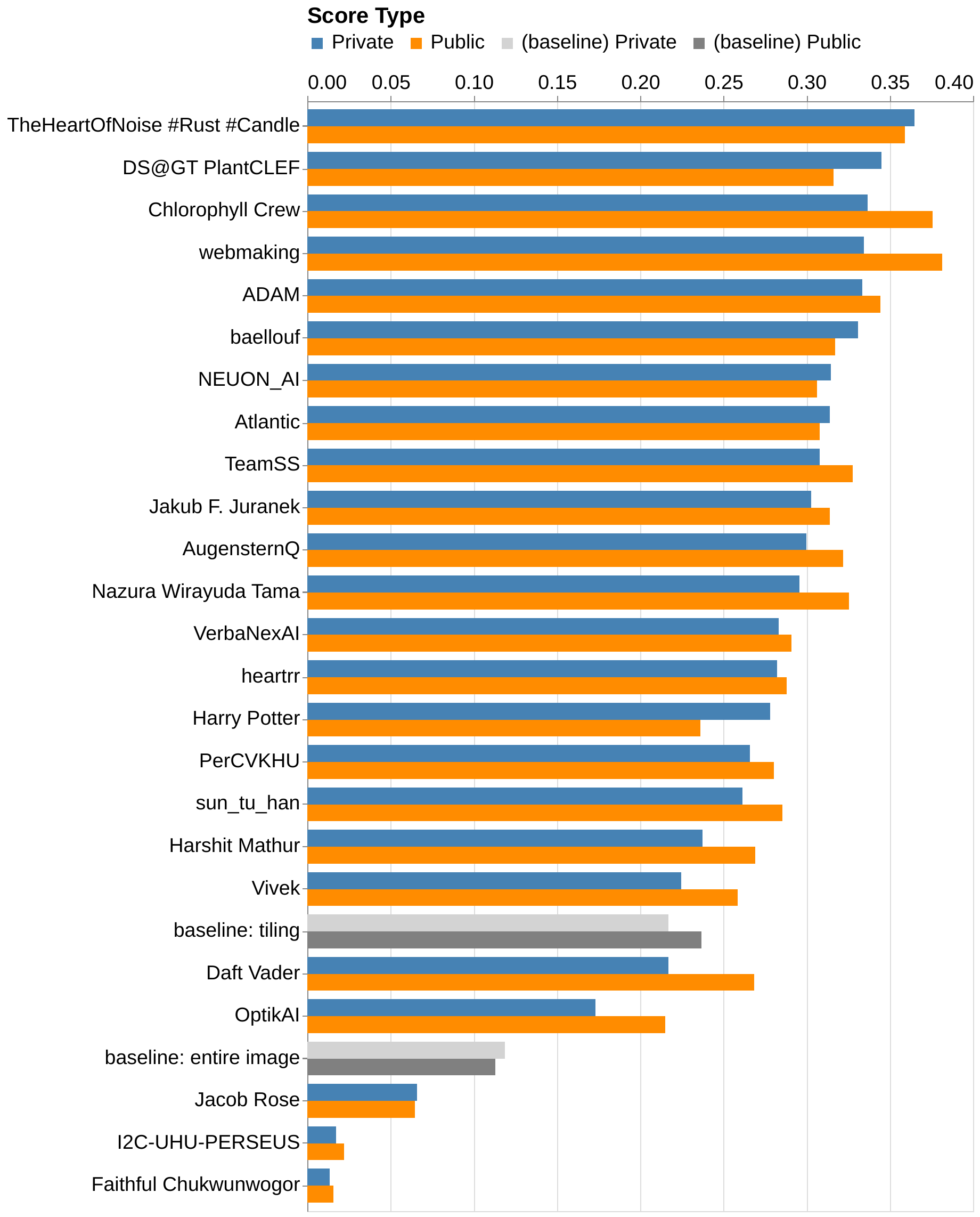}
    \caption{Best run for each team; Mean Sample-Averaged F1 Score Per Transect (each transect includes multiple quadrats from the same site).}
    \label{fig:PlantCLEF2025Scores}
\end{figure}

Figure~\ref{fig:PlantCLEF2025Scores} presents the best performance achieved by each team, measured as the highest private macro-averaged F1 score per sample. As shown, the task remains highly challenging, with the leading submission attaining a macro-averaged F1 score of just 0.3648 on the private leaderboard. Nevertheless, this result marks a significant improvement over the previous edition, where the highest score reached just 0.2873, reflecting notable methodological progress over the past year. The leaderboard shows a wide dispersion of scores, ranging from approximately 0.37 down to 0.01, illustrating the varying effectiveness of the proposed approaches. Comparing the public and private leaderboard rankings reveals substantial differences in final standings. For instance, the team TheHeartOfNoise moved from 3rd place on the public leaderboard to 1st on the private one, webmaking dropped from 1st to 4th, while DS@GT PlantCLEF rose from 9th to 2nd place. Several other teams also experienced notable rank shifts within the top 10, indicating a hyper-parameter tuning overfitting problem. In comparison, the official tiling-based baseline achieved a macro-averaged F1 score of 0.21708, a performance surpassed by most of participating teams.

\section{Discussion}

The main outcomes we can derive from those results are the following:\\

\textbf{Score interpretation:} The best score remains below 0.37 in sample-averaged F1 but shows clear progress compared to last year’s result (0.29). Considering that there are about 8 species per quadrat on average, this score typically reflects 4–5 correct predictions per image, with relatively few false positives, indicating a good trade-off between precision and recall under challenging conditions.\\

\textbf{Overfitting and generalization:} The webmaking team, despite achieving the highest public score after over 200 submissions, dropped to 4th place on the private leaderboard. In contrast, DS@GT PlantCLEF, ranked only 9th publicly, rose to 2nd. This suggests classic leaderboard overfitting and highlights the challenge of generalization across unseen test data.\\

\textbf{Impact of image preprocessing:} The top-ranked submission highlights the importance of aligning test-time image preprocessing with the properties of the training data. In particular, JPEG recompression of quadrat images using 4:2:2 chroma subsampling—matching the dominant YCrCb format of the single-species training set—proved beneficial. Combined with Lanczos resizing, this strategy reduced the distribution gap between training and test data, effectively improving generalization. Such low-level operations, often underestimated, can yield substantial performance gains by mitigating domain shift.\\

\textbf{Tiling strategies:} Most participants likely explored various tiling strategies, experimenting with different tile sizes, overlap settings, fusion or decision mechanisms, among other configurations. While each team identified an optimal configuration tailored to their own method, there is no clear consensus yet on a universally best tiling approach. The close proximity of the top-performing scores suggests that multiple strategies can yield comparable performance under the current benchmark conditions.\\

\textbf{Aggregation strategies:} Several teams emphasized post-hoc aggregation as a key performance lever. NEUON\_AI obtained one of the best scores without any model fine-tuning, relying solely on Bayesian Model Averaging to combine patch-level predictions using confidence metrics (variance, entropy, etc.). This confirms that carefully designed decision-level aggregation mechanisms can rival more complex model-centric strategies.\\

\textbf{Non-vegetative region filtering:} Several teams attempted to improve predictions by isolating vegetative regions. ADAM explicitly restricted classification to plant-relevant zones identified via a prototype-reconstruction mechanism, effectively filtering out non-vegetative areas. Webmaking applied a softer down-weighting approach, using zero-shot object segmentation to reduce the influence of non-plant regions, without enforcing a strict rejection. DS@GT also explored rejection strategies, but reported no measurable improvement. Overall, while intuitively relevant, vegetation-focused filtering did not consistently yield significant performance gains, and this hypothesis was not clearly supported by internal ablation studies reported in the working notes.\\

\textbf{Limitations of object detection for vegetation filtering:} The I2C-UHU-PERSEUS team used YOLOv11 to isolate vegetative regions before classifying them with InceptionV3. This approach faces two major limitations in the PlantCLEF setting: first, YOLO struggles to detect discrete plant instances in high-resolution plots with dense, overlapping vegetation and no clear object boundaries; second, using InceptionV3 instead of the ViT model provided by the organizers likely limited classification performance. Overall, object detection as a pre-filtering stage appears poorly suited to the structural complexity and fine-grained nature of the task.\\

\textbf{Use of taxonomic hierarchy:} Chlorophyll Crew incorporated multi-head classifiers for species, genus, and family, assuming that taxonomic relations would improve representation learning. This hypothesis appears validated, as their method ranked among the top submissions, suggesting that hierarchical supervision may help disambiguate difficult cases.\\

\textbf{Ecological priors and context:} Webmaking integrated ecological post-processing using GBIF occurrence data and Ellenberg indicator values niche similarity.These priors enhanced F1 scores by almost 0.01 without any model retraining, showing the potential of external ecological knowledge to complement visual inference.\\

\textbf{Minimal benefit from SSL pretext tasks:} NEUON\_AI tested several self-supervised learning objectives to reduce domain shift but no improvement over the baseline model was observed. This suggests that, when fine-tuning data is limited, naive SSL approaches are likely to offer limited utility, unless pretext tasks are carefully adapted to the target domain.\\

\textbf{The highest purely visual performance}, i.e. the one without the use of metadata or rejection mechanisms, is 0.35016 obtained by TheHeartOfNoise, with test-time image preprocessing and multi-scale tiling with nine different sizes. Further improvements are obtained grouping images by transect and with manual optimization.

\section{Conclusion}

The PlantCLEF 2025 results confirm the structural difficulty of the task, as observed in the previous edition. Despite a clear improvement in the sample-averaged F1 score (from 0.29 in 2024 to 0.36 this year), the overall performance remains modest given the complexity of the multi-label, fine-grained classification on high-resolution images with a significant domain shift.

Several trends can be observed. Low-level preprocessing operations, such as JPEG recompression with appropriate chroma subsampling and resizing aligned with training conditions, were shown to significantly improve generalisation performance. In contrast, strategies focused on isolating vegetative regions did not produce consistent measurable gains and appear to be highly implementation-dependent.

Structural and ecological priors were rarely exploited. Taxonomic hierarchies and external occurrence or niche data were only marginally used, despite their potential to filter false positives. The complementary pseudo-quadrat dataset provided for domain adaptation remained entirely unused by any of the teams that submitted working notes.

Finally, large rank shifts between public and private leaderboards illustrate persistent overfitting issues. Performance gains were primarily obtained through incremental refinement of established pipelines—tiling, aggregation, and filtering — rather than through paradigm shifts or architectural innovations.

From a methodological perspective, promising avenues include multi-label models capable of handling full-resolution images in a single pass or through end-to-end tiling mechanisms. Ideally, such approaches should be robust to low-level encoding variations and designed to integrate ecological or structural priors. Regarding the challenge, the results confirm that significant potential remains to be explored, both in terms of architecture and domain adaptation. The growing number of participants this year suggests an increasing interest in the task, likely driven by greater visibility via Kaggle and its relevance to biodiversity monitoring - an area that addresses a complex domain shift problem with considerable societal value.

\begin{acknowledgments}

The research described in this paper was partly funded by the European Commission via the GUARDEN and MAMBO projects, which have received funding from the European Union’s Horizon Europe research and innovation program under grant agreements 101060693 and 101060639. The opinions expressed in this work are those of the authors and are not necessarily those of the GUARDEN or MAMBO partners or the European Commission.

This project was provided with computer and storage resources by GENCI at IDRIS thanks to the grant 2024-AD010113641R2 on the supercomputer Jean Zay's the V100 partition.

\end{acknowledgments}

\section*{Declaration on Generative AI}
During the preparation of this work, the author(s) used ChatGPT-4o and Deepl in order to: Grammar and spelling check. After using these tool(s)/service(s), the author(s) reviewed and edited the content as needed and take(s) full responsibility for the publication’s content.


\end{document}